\pdfoutput=1

\documentclass[11pt]{article}

\usepackage{acl}

\usepackage{times}
\usepackage{latexsym}

\usepackage[T1]{fontenc}

\usepackage[utf8]{inputenc}

\usepackage{microtype}

\usepackage{makecell}
\usepackage{times}
\usepackage{latexsym}
\usepackage{tabularx}
\usepackage{array}
\usepackage{multirow}
\usepackage{siunitx}
\usepackage{booktabs}
\usepackage[pdftex]{graphicx}
\usepackage{enumitem}
\usepackage{xspace}
\usepackage{caption}

\usepackage{url}            
\usepackage{amsmath,amsthm,amsfonts,amssymb,bm,stmaryrd}       
\usepackage{nicefrac}       
\usepackage{enumitem}
\usepackage[noend]{algpseudocode}
\usepackage{algorithm}
\usepackage{mathtools}
\usepackage{nccmath}
\usepackage{multirow}
\usepackage{bigdelim}
\usepackage{color, colortbl}
\usepackage{xcolor}		
\definecolor{darkblue}{rgb}{0, 0, 0.5}
\hypersetup{colorlinks=true, citecolor=darkblue, linkcolor=darkblue, urlcolor=darkblue}
\usepackage{booktabs}
\usepackage{dcolumn}
\newcolumntype{d}{D{.}{.}{-1}}
\newcolumntype{z}{D{(}{\ (}{1.1}}
\usepackage{graphicx}
\usepackage{titlesec}
\usepackage{hyperref}
\usepackage{url}
\usepackage{framed}
\usepackage{tcolorbox}
\usepackage{subcaption}
\usepackage{graphicx}
\usepackage[normalem]{ulem}
\useunder{\uline}{\ul}{}
\usepackage{fontawesome5}
\usepackage{calc}

\usepackage{cleveref}

\crefformat{section}{\textcolor{blue}{\S#2#1#3}} 
\crefformat{subsection}{\textcolor{blue}{\S#2#1#3}}
\crefformat{subsubsection}{\textcolor{blue}{\S#2#1#3}}
\crefformat{equation}{\textcolor{orange}{Eq.~(#1)}}

\makeatletter
\def\blfootnote{\xdef\@thefnmark{}\@footnotetext}
\makeatother

\newcommand{\datasetname}{\textsc{RefNLI}\xspace}

\newcommand\pennlogo{\raisebox{-2pt}{\includegraphics[width=0.7em]{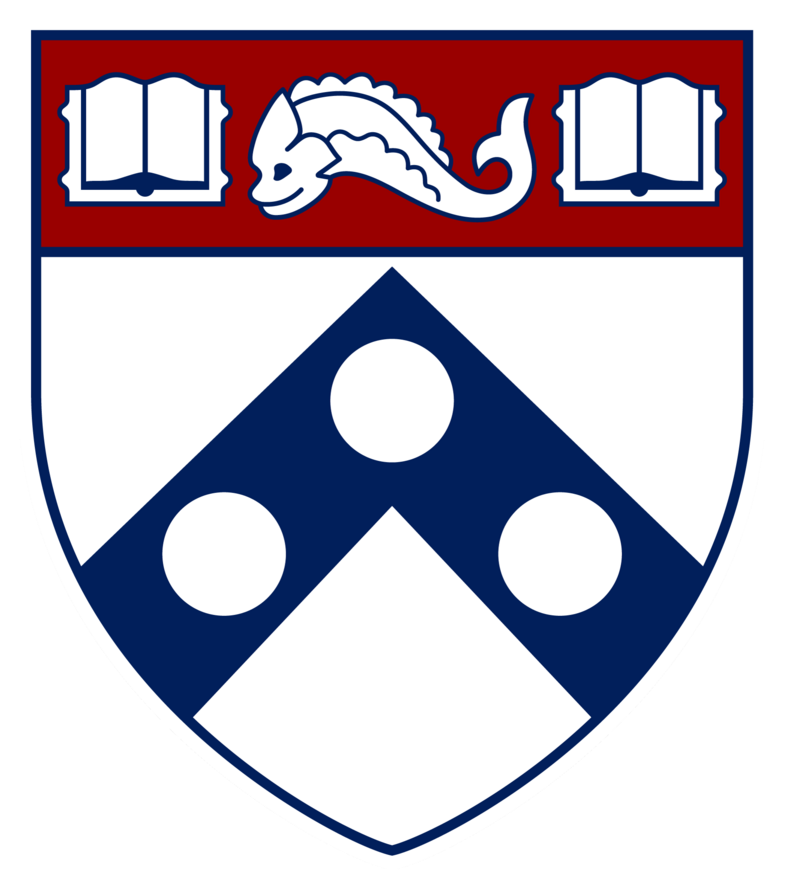}}}
\newcommand\gdmlogo{\raisebox{-2pt}{\includegraphics[width=0.9em]{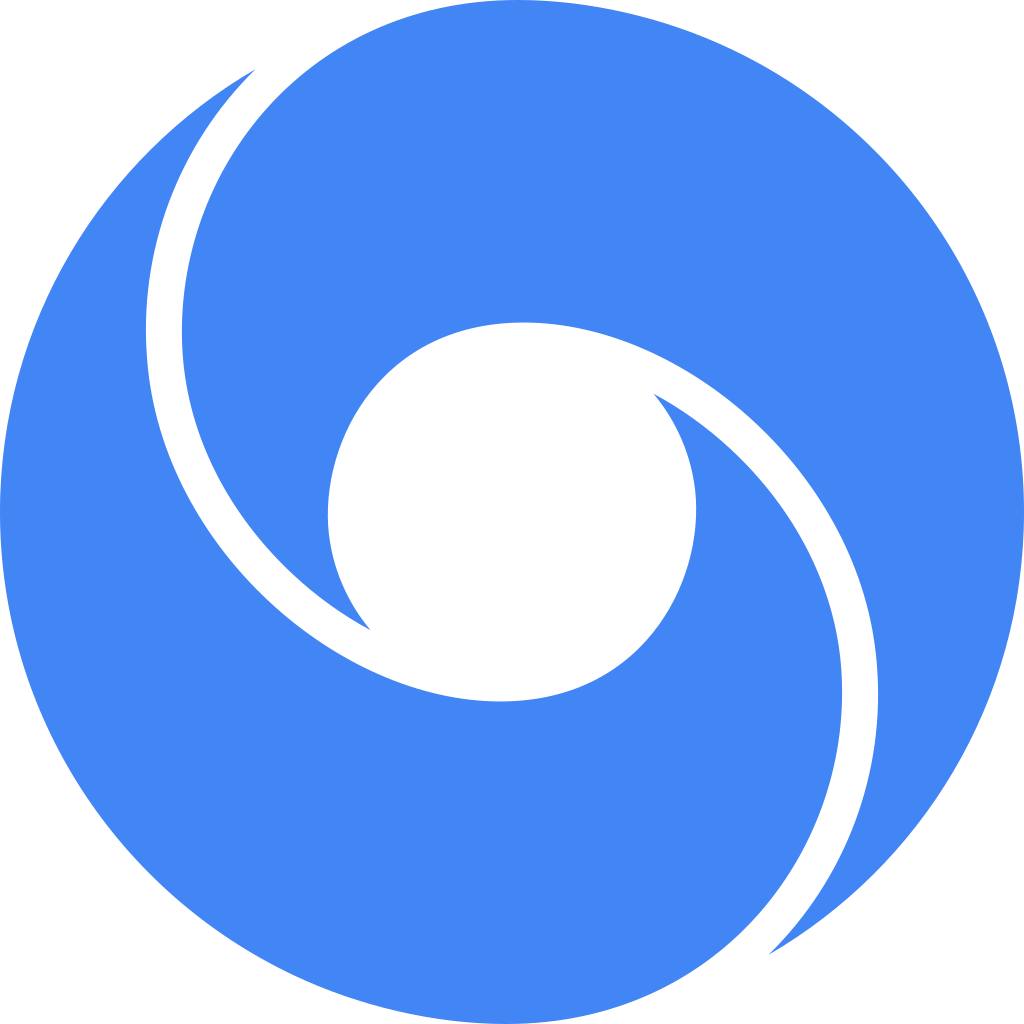}}}
\newcommand\googlogo{\raisebox{-2pt}{\includegraphics[width=0.9em]{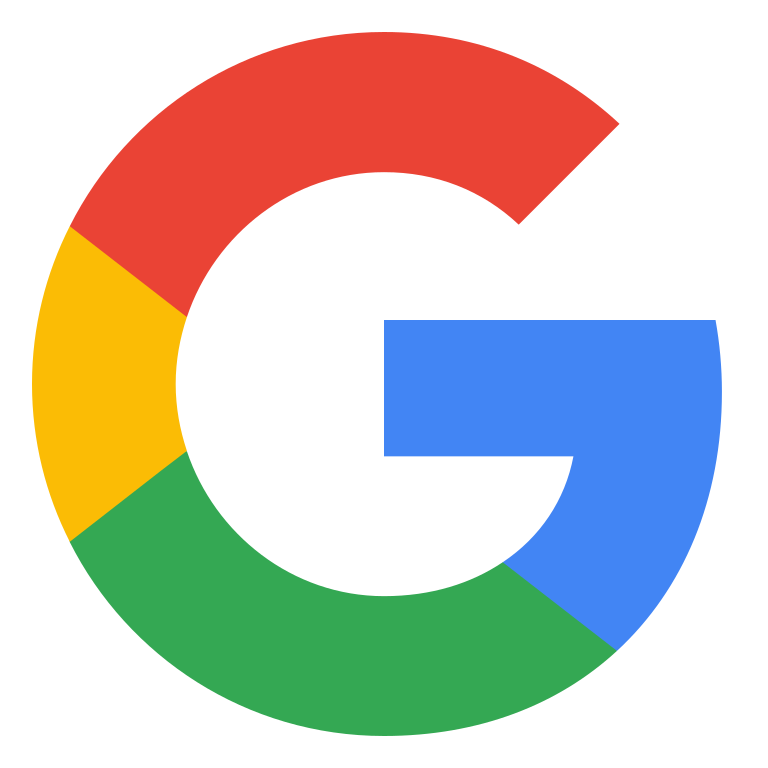}}}
\newcommand\msftlogo{\raisebox{-2pt}{\includegraphics[width=0.9em]{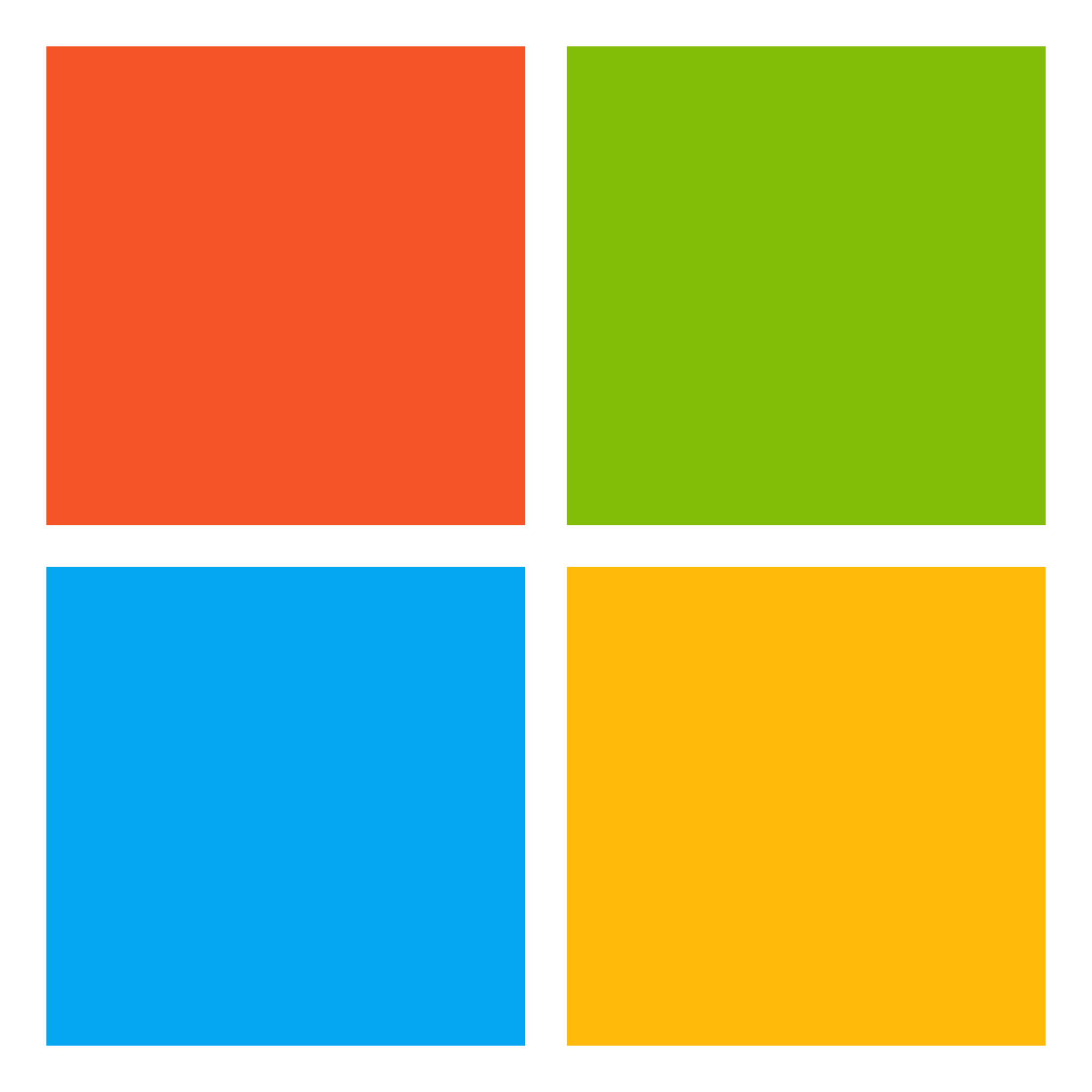}}}

\title{On Reference \textit{(In-)}Determinacy in Natural Language Inference}

\author{Sihao Chen\textsuperscript{\msftlogo}\thanks{\,~Work done during Sihao's and Chaitanya's internship at Google. Sihao was a Ph.D. student at the University of Pennsylvania at the time.  } \, \textbf{Chaitanya Malaviya}\textsuperscript{\pennlogo} \, \textbf{Alex Fabrikant}\textsuperscript{\gdmlogo} \\ \textbf{Hagai Taitelbaum}\textsuperscript{\googlogo} \, \textbf{Tal Schuster}\textsuperscript{\gdmlogo} \, \textbf{Senaka Buthpitiya}\textsuperscript{\gdmlogo} \, \textbf{Dan Roth}\textsuperscript{\pennlogo} \vspace{0.1in}
\\
\textsuperscript{\pennlogo} University of Pennsylvania \,
\textsuperscript{\gdmlogo} Google DeepMind \\
\textsuperscript{\googlogo} Google Research \, \textsuperscript{\msftlogo} Microsoft \\
}

\begin{document}
\maketitle

\begin{abstract}
We revisit the \textit{reference determinacy} (RD) assumption in the task of natural language inference (NLI), i.e., the premise and hypothesis are assumed to refer to the same context when human raters annotate a label. 
While RD is a practical assumption for constructing a new NLI dataset, we observe that current NLI models---which are typically trained solely on hypothesis-premise pairs created with the RD assumption---fail in downstream applications such as fact verification, where the input premise and hypothesis may refer to different contexts. To highlight the impact of this phenomenon in real-world use cases, we introduce \datasetname, a diagnostic benchmark for identifying reference ambiguity in NLI examples.  
In \datasetname, the premise is retrieved from a knowledge source (i.e. Wikipedia) and does not necessarily refer to the same context as the hypothesis. With \datasetname, we demonstrate that finetuned NLI models and few-shot prompted LLMs both fail to recognize context mismatch, leading to $>80\%$ false contradiction and $>50\%$ entailment predictions. We discover that the existence of reference ambiguity in NLI examples can in part explain the inherent human disagreements in NLI, and provide insight into how the RD assumption impacts NLI dataset creation process.
\\\\
\faGithub \,\,\url{https://github.com/refnli-authors/refnli}


\end{abstract}

\section{Introduction}
Natural Language Inference (NLI), or Recognizing Textual Entailment (RTE), provides a general task format for evaluating the semantic relation between two pieces of text, where a system is expected to predict if a hypothesis statement can be inferred from a given premise.
For the past few decades, NLI has been the centerpiece for the development and evaluation of language understanding systems \cite{dagan2005pascal, bowman-etal-2015-large, williams-etal-2018-broad, nie-etal-2020-adversarial}.  

As the use of NLI now spreads across a wider variety of downstream applications, such as text classification \cite{yin-etal-2019-benchmarking}, fact verification \cite{schuster-etal-2021-get}, hallucination detection \cite{kryscinski-etal-2020-evaluating}, text attribution \cite{gao-etal-2023-rarr}, etc., 
it is important to understand how the \textit{definitions} and \textit{assumptions} made for collection of previous NLI datasets and models trained on them affect their usefulness in downstream use cases.

\begin{table}[t]
\centering
    \begin{tabular}{p{0.95\linewidth}}
        \toprule
        \textbf{Premise}: A black race car starts up in front of a crowd of people.	  \\
        \textbf{Hypothesis}: A man is driving down a lonely road.  \\
        \midrule
        \textbf{Original Label in SNLI}: \textcolor{red}{Contradiction} (5/5) \\
        \textbf{Label} {\small (without \textit{Reference Determinacy})}: \textcolor{blue}{Neutral} \\
        \bottomrule
    \end{tabular}
    \caption{An example from the SNLI dataset \cite{bowman-etal-2015-large} with all five annotators agreeing on the hypothesis contradicting the premise, under the \textit{reference determinacy} assumption, i.e.\ the events described in the premise and hypothesis happen on the same road. Without the assumption, the label would likely be neutral.}
    \vspace{-10pt}
    \label{tab:lead}
\end{table}

In this paper, we revisit and study the effect of \textit{reference determinacy} (RD), a common assumption formed in the labeling of NLI datasets.
With RD, the NLI label between a pair of premise and hypothesis is annotated under the assumption that the pair refer to the same context \cite{bowman-etal-2015-large}.
We illustrate the idea behind RD through an example in \autoref{tab:lead}, where the premise and the hypothesis describe two different events. 
The premise \textit{contradicts} the hypothesis (i.e.,\ premise $\rightarrow \neg$ hypothesis) only when we opt to assume that the two events happen on the same road at the same time. Otherwise, the pair would be labeled \textit{neutral}, as the two events are most likely unrelated. 

RD is a practical assumption for the NLI label definition. Without the RD assumption, the entailment and contradiction relations would only exist when the hypothesis and premise describe functional relations that are universally true or false \cite{ritter-etal-2008-contradiction}, e.g. factual knowledge about an entity. For this reason, most large-scale NLI benchmarks follow the RD assumption during their annotation processes (\cref{ssec:prelim-rd}). However, if we train NLI models exclusively on hypothesis-premise pairs created with the RD assumption, this could lead to the resulting models having limited ability to recognize if a hypothesis is relevant to a premise.

We demonstrate the trickle-down effects of such NLI model behavior in downstream tasks such as fact verification.
Specifically, we sample claims from FEVER \cite{thorne-etal-2018-fever} and VitaminC \cite{schuster-etal-2021-get} and study how NLI models behave when used to verify against evidence retrieved from the web. From the sampled claims, we construct the \datasetname benchmark (\cref{sec:benchmark}), which features 1,143 NLI pairs with expert judgements for whether the premise and hypothesis refer to the same context, as well as the correct NLI label.

With \datasetname, we observe that both finetuned NLI models as well as LLMs few-shot prompted to classify 3-way NLI labels often fail to recognize context mismatches, which leads to many false entailment and contradiction predictions. On five popular NLI datasets (\cref{sec:results}), we demonstrate that different combinations of training datasets result in similar type of reference (in-)determinacy problem in the finetuned model. This indicates the existence of a reference determinacy bias in all five datasets, which we discuss in the context of how each of the five datasets are created. We propose strategies to filter out entailment or contradiction examples labeled only due to the reference determinacy assumption, and show this can mitigate the reference determinacy bias of finetuned NLI models at inference time.

Reference determinacy, we discover, can also partly explain part the distribution of human disagreements of NLI labels, a problem known to be widespread in popular NLI datasets  \cite{pavlick-kwiatkowski-2019-inherent,nie-etal-2020-learn}. Our analysis shows that human typically disagree more on examples where reference determinacy cannot be safely assumed, and disagreements happen when annotators are instructed to do so regardless.

In summary, our contributions in the paper are: 
\begin{itemize}[leftmargin=*, itemsep=0em]
    \item We introduce the \datasetname benchmark, a dataset featuring 1,143 examples for studying the the effect of reference determinacy in NLI, a common assumption in the creation processes of NLI datasets.  
    \item With \datasetname, we investigate the downstream impact of the reference determinacy assumption of NLI dataset creation process. We show that finetuned NLI models and LLMs exhibit reference determinacy bias and often fail to recognize context mismatches.
    \item We discover and study the connection of the reference determinacy assumption to the inherent human disagreement on NLI labels.  
\end{itemize}

\section{The Reference Determinacy Assumption}
\label{ssec:prelim-rd}
When we create and label NLI examples, \textit{reference determinacy} (RD) is a practical assumption for guaranteeing the correctness and consistency of annotated labels. For instance, suppose a hypothesis and premise pair both mention \textit{John Doe}, the perceived entailment or contradiction relation could change based on whether we believe the two ``John Doe''s are a single real-world person. 

\paragraph{The creation processes of most NLI datasets assume reference determinacy.} 

For example, in SNLI \cite{bowman-etal-2015-large} and MNLI \cite{williams-etal-2018-broad}, annotators were asked to write novel hypotheses that are either true/false/neutral in the context of a given premise. During  labeling, the hypothesis is interpreted in the context of the premise, where entities and events in the two are assumed to be co-refer between the hypothesis and premise
As a result, we see examples like in \autoref{tab:lead}, where majority of the annotators would agree on the contradiction or entailment label, when the premise and hypothesis likely refer to different events without the RD assumption.  

Following
MNLI and SNLI, large-scale NLI datasets, e.g. \citet{marelli-etal-2014-sick, Khot2018SciTaiLAT, conneau-etal-2018-xnli}, among others,
typically use similar processes to create and label hypotheses from given premises. 
Here, we study models trained on MNLI, SNLI, plus other notable datasets including ANLI \cite{nie-etal-2020-adversarial} and VitaminC \cite{schuster-etal-2021-get}. We aim to understand the behavior of models trained on these datasets at recognizing relevance between hypothesis and premise pairs.




\begin{table*}[t]
    \centering
\small
\begin{tabular}{p{0.2\linewidth}p{0.45\linewidth}cc}
\toprule
\multicolumn{1}{c}{\textbf{Hypothesis}} &
\multicolumn{1}{c}{\textbf{Premise}} &
\multicolumn{1}{c}{\textbf{RefNLI Label}} &
\multicolumn{1}{c}{\textbf{Model Pred.}}

\\ \midrule
Sabbir Khan made his directorial debut in 2001. &
In 2009 he made his directorial debut with the film ``Kambakkht Ishq'' (2009) that starred Akshay Kumar and Kareena Kapoor. &
\textcolor{orange}{Ambiguous} & 
\textcolor{magenta}{Contradiction}
\\ \midrule
\multicolumn{4}{p{0.95\linewidth}}{\textcolor{gray}{\textbf{Explanation}: The premise contains the ambiguous reference ``he'' as the director that made the debut. However, there exists an assignment of the pronoun "he" such that the hypothesis can be contradicted. In this case, without resolving the pronoun reference, the NLI label can not be determined. Therefore, the label here is ``Ambiguous''. }}
\\ \midrule
Wales has a large region rich in coal deposits. &
Recent explorations have revealed prospective deposits of rare-earth elements, a company is proposing further analysis of these mineral deposits. &
Neutral & 
\textcolor{magenta}{Contradiction}
\\ \midrule
\multicolumn{4}{p{0.95\linewidth}}{\textcolor{gray}{\textbf{Explanation}: The premise does not specify the location of the deposits of rare-earth elements. However, as coal is not a type of rare-earth element, we know for sure that whichever location the premise is referring to, the premise here cannot be used to support or contradict the hypothesis. Therefore, the label is ``Neutral''. }}
\\ \midrule
Same Old Love is a work of music. &
``Same Old Love'' was also performed on ``The Ellen DeGeneres Show'', ``The Tonight Show Starring Jimmy Fallon'', 2015 American Music Awards, and at the 2015 Billboard Women in Music. &
\textcolor{teal}{Entailment} & 
\textcolor{magenta}{Contradiction}
\\ \midrule
\multicolumn{4}{p{0.95\linewidth}}{\textcolor{gray}{\textbf{Explanation}: The annotators agree that it's reasonable to assume that ``Same Old Love'' refers to the same thing without ambiguity here. Therefore, the label here is ``Entailment''. }}
\\ \midrule
Buffy the Vampire Slayer is exclusively a Japanese television series. &
``Buffy the Vampire Slayer'' comics refer to comic books based on the television series ``Buffy the Vampire Slayer'' & \textcolor{magenta}{Contradiction} &
\textcolor{teal}{Entailment}
\\ \midrule
\multicolumn{4}{p{0.95\linewidth}}{\textcolor{gray}{\textbf{Explanation}: Even though there could be a Japanese television series named Buffy the Vampire Slayer, the premise would refute the hypothesis that it is exclusively a Japanese television series. Therefore, the label here is ``Contradiction''. }}

\\ \bottomrule
\end{tabular}%
\caption{
Examples from our study and the \datasetname benchmark. Compared to the usual three-way NLI label set, i.e. \emph{entailment}, \emph{neutral} and \emph{contradiction}, we explicitly distinguish the \emph{ambiguous} cases, where reference determinacy between the hypothesis and premise is meaningful yet cannot be established. ``Model Pred.'' shows predictions made by the RoBERTa-based NLI model \citet{nie-etal-2020-adversarial} under three-way classification. 
}
\label{tab:refnli-examples}
\end{table*}

\section{A Case Study of Reference (In-)Determinacy}
\label{sec:benchmark}
NLI models are typically finetuned exclusively on examples created with the reference determinacy assumption. We first study the effect of the RD assumption when we use such NLI models to solve downstream tasks.
Specifically, we aim to understand how an NLI model would behave in a realistic scenario where the premise can be irrelevant to the hypothesis.
In such cases, if there exists enough information in the evidence to establish reference determinacy, i.e. humans would be able to determine whether the evidence is related to the claim or not, an ideal NLI model should be able to correctly derive the NLI label. 

Motivated by this, we study the use of NLI for the task of fact verification. 
We construct the \datasetname benchmark, which features 1,143 pairs of claim and retrieved Wikipedia evidence sentence, with human-labeled reference determinacy and entailment relations.
 



\subsection{Sampling Claims and Evidence}
We start by sampling claims from the validation and test splits of FEVER \cite{thorne-etal-2018-fever} and VitaminC \cite{schuster-etal-2021-get}. With each claim, we use BM25 to retrieve the top-10 passages from an English Wikipedia dump from 2018-07-01 with \texttt{pyserini} \cite{Lin_etal_SIGIR2021_Pyserini}  .\footnote{\url{https://github.com/castorini/pyserini}} Note that most of the retrieved passages would not be related to the entity or event described in the claim.  
Next, given each claim and each sentence in the top-10 retrieved passages, 
we classify their relation with a widely-used, pretrained RoBERTa model \cite{liu2019roberta} finetuned on a mixture of NLI datasets from \citet{nie-etal-2020-adversarial}.

\begin{figure}
    \centering
    \includegraphics[width=0.9\linewidth]{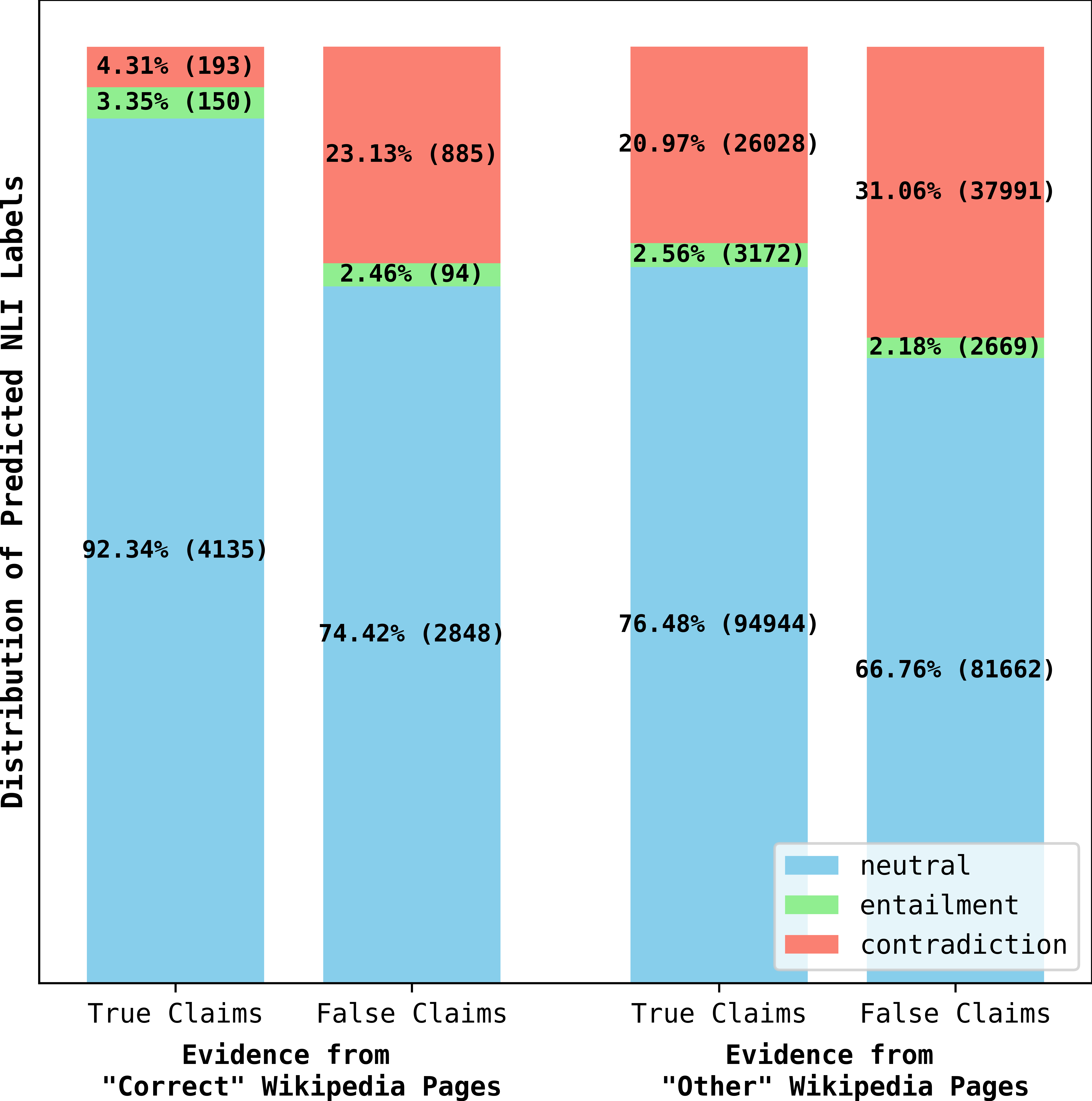}
    \caption{The distribution of label predictions by RoBERTa NLI mixture model from  \citet{nie-etal-2020-adversarial} when used to verify claims against retrieved evidence sentences from the correct vs. (most likely) irrelevant Wikipedia pages.}
    \label{fig:fever-dist}
\end{figure}

\paragraph{NLI model predicts many false contradictions.} 
On the development set of FEVER, we compare how the model behaves when a claim is verified against evidence sentences from the ``correct'' Wikipedia page, as labeled in FEVER, compared to sentences from other Wikipedia pages, which are likely to be irrelevant to the claim. In the later case, we expect the NLI model to discover very little \textit{supporting} or \textit{contradicting} evidence, as the page is unlikely to be relevant to the claim.

\autoref{fig:fever-dist} shows the distribution of the NLI model's label predictions when used to verify claims labeled as \textit{supported (True)} or \textit{refuted (False)} by Wikipedia in FEVER. 
We observe that apart from the case where true claims are verified against sentences from the correct Wikipedia page, NLI models make contradiction predictions much more frequently than entailments in all the other three cases. 
While finding contradictions of false claim in the \textit{correct}  Wikipedia page where the refuting evidence comes from is what we want to see, interestingly we observe that the NLI model predict much more contradictions against \emph{irrelevant} Wikipedia pages, i.e. pages about a different entity. 
In cases where the sentence comes from such irrelevant Wikipedia pages, the pattern of potential ``false contradictions'' from the model is largely visible. 
The finding here echoes our initial hypothesis, suggesting the NLI model seems to be lacking the ability to recognize whether an evidence sentence refers to the same context as the claim.


\subsection{The \datasetname Benchmark}
\label{ssec:benchmark}
To further validate our hypothesis and understand why NLI models behave this way, we design a human study and analyze the example predictions made by NLI models in this setting.

From the set of examples where the RoBERTa NLI model predicts entailment or contradictions, and the evidence does not come from the correct Wikipedia page, we sample a subset for human annotation uniformly at random. The authors of the paper then annotate each claim and evidence sentence pair with one of the following four labels. The label set here follows what is expected from a fact verification system, when asked to verify a given claim in the context of the evidence.

\begin{itemize}[leftmargin=*, itemsep=0em]
    \item \emph{Entailment}: if the human annotator thinks that the evidence and claim likely refer to the same context, and the evidence is sufficient to fully support the claim.
    \item \emph{Contradiction}: if the human annotator thinks that the evidence and claim likely refer to the same context, and claim is unlikely to be true given the evidence.
    \item \emph{Ambiguous}: if it is unclear whether the claim and the evidence refer to the same context (e.g. contain ambiguous reference), and there exist multiple possible assignments or interpretations of references that could make the example fall into at least 2 of the other 3 labels.
    \item \emph{Neutral}: if it is clear that the evidence cannot support or contradict the claim in any way, i.e. there exists no interpretation or assignment of references of the evidence where it can support or contradict the claim.
\end{itemize}

Compared to the usual 3-way NLI labels, the label set here is designed to distinguish where reference determinacy cannot be safely established between a hypothesis and a premise. 
When there exist ambiguous references, a fact verification system should not make any assumption about the reference and conclude its entailment relation with the evidence. 
Note that even if there exist ambiguous references, as long as the premise is unrelated to the hypothesis, no matter how the ambiguous reference is interpreted, the system could still deem the claim as \emph{neutral}, as there is no way that the claim can be supported or refuted by the evidence. This follows the intuition that ambiguity in reference determinacy only matters when there exists an interpretation where the evidence could be related to the claim.
To help understand the motivation behind the label set design, we include one example of each label in \autoref{tab:refnli-examples}. For instance, the first example is labeled as ambiguous, as there exist an possible assignment of the pronoun \emph{he} $\rightarrow$ \emph{Sabbir Khan}, such that a fact verification should not conclude that the hypothesis is irrelevant to the evidence. On the other hand, in the second example of ``Wales coal deposit'', as Wales coal deposit is not a type of rare-earth element, the premise is always going to be irrelevant to the claim, no matter what the assignments of the ambiguous references in the premise are. 
We include a more detailed description of the annotation guidelines and discussion of corner cases in Appendix \ref{appendix:rater-guidelines}.

\paragraph{The difference between \textit{neutral} vs. \textit{ambiguous}.}  From the NLI task's perspective, the notable difference is that \emph{neutral} hypothesis-premise pairs themselves contain enough information for humans to judge that the premise is irrelevant to the claim. In such cases, it is reasonable to expect a good NLI model to make the correct prediction, whereas for \emph{ambiguous} examples, the correct label cannot be determined without the RD assumption. In our study, we do not expect NLI models to work well for ambiguous examples. NLI models' behavior with respect to ambiguity is investigated in greater detail in a recent study from \citet{liu-etal-2023-afraid}. 

\begin{table*}[t]
\centering \small
\resizebox{\textwidth}{!}{%
\begin{tabular}{|c|ccc|ccc|ccc|}
\toprule
 Training Data & \multicolumn{3}{c|}{Contradiction (66)} & \multicolumn{3}{c|}{Neutral (905)} & \multicolumn{3}{c|}{Entailment (37)} \\
 (Model) & Precision & Recall & AUROC & Precision & Recall & AUROC  & Precision & Recall & AUROC  \\
\midrule
\texttt{ALL} &  15.76\scriptsize{$\pm$3.74} & 92.42\scriptsize{$\pm$6.73} & 90.91 & 98.99\scriptsize{$\pm$0.92} &  53.92\scriptsize{$\pm$3.36} & 87.49 & 25.78\scriptsize{$\pm$7.92} & 89.18\scriptsize{$\pm$10.75} & 94.53\\
\midrule
\texttt{ALL} - SNLI & 17.39\scriptsize{$\pm$4.04} & 90.91\scriptsize{$\pm$7.18} & 90.51 & 98.80\scriptsize{$\pm$0.88} & 63.75\scriptsize{$\pm$3.23} & 89.25 & 36.71\scriptsize{$\pm$10.95} & 78.37\scriptsize{$\pm$13.56} & 91.74\\
\texttt{ALL} - MNLI & 22.30\scriptsize{$\pm$5.23} & 90.91\scriptsize{$\pm$7.15} & 89.62 & 98.93\scriptsize{$\pm$0.80} & 71.93\scriptsize{$\pm$3.09} & 89.42 & 38.27\scriptsize{$\pm$10.43} & 83.78\scriptsize{$\pm$12.19} & 94.68 \\
\texttt{ALL} - ANLI & 19.87\scriptsize{$\pm$4.58} & 89.39\scriptsize{$\pm$7.61} & 88.50 & 98.46\scriptsize{$\pm$1.00} & 70.60\scriptsize{$\pm$3.09} & 88.02 & 50.00\scriptsize{$\pm$12.83} & 83.78\scriptsize{$\pm$12.33} & 93.92 \\
\texttt{ALL} - Fever(NLI) & 15.71\scriptsize{$\pm$3.58} & 90.81\scriptsize{$\pm$7.14} & 87.85 & 98.35\scriptsize{$\pm$1.11} & 59.44\scriptsize{$\pm$3.28} & 84.47 & 37.97\scriptsize{$\pm$11.23} & 81.08\scriptsize{$\pm$13.32} & 96.13 \\
\texttt{ALL} - VitaminC & 14.06\scriptsize{$\pm$3.39} & 92.42\scriptsize{$\pm$6.76} & 88.37 & 98.85\scriptsize{$\pm$10.35} & 47.40\scriptsize{$\pm$3.31} & 83.68 & 23.57\scriptsize{$\pm$6.99} & 89.19\scriptsize{$\pm$10.73} & 94.57 \\
\midrule
SNLI & 8.40\scriptsize{$\pm$2.08} & 93.94\scriptsize{$\pm$5.93} & 72.21 & 97.07\scriptsize{$\pm$2.33} & 21.99\scriptsize{$\pm$2.71} & 66.46 & 50.77\scriptsize{$\pm$12.16} & 89.18\scriptsize{$\pm$9.85} & 96.70 \\
MNLI & 10.91\scriptsize{$\pm$3.69} & 93.94\scriptsize{$\pm$8.04} & 88.48 & 98.20\scriptsize{$\pm$0.93} & 42.21\scriptsize{$\pm$2.32} & 81.17 & 62.75\scriptsize{$\pm$9.57} & 86.48\scriptsize{$\pm$7.88} & 94.93 \\
ANLI & 19.04\scriptsize{$\pm$4.52} & 90.91\scriptsize{$\pm$7.19} & 92.18 & 98.85\scriptsize{$\pm$0.86} & 66.96\scriptsize{$\pm$3.12} & 91.66 & 38.75\scriptsize{$\pm$10.91} & 83.78\scriptsize{$\pm$12.03} & 95.60\\
Fever(NLI) & 6.29\scriptsize{$\pm$4.08} & 13.64\scriptsize{$\pm$8.90} & 58.04 & 90.42\scriptsize{$\pm$2.46} & 66.74\scriptsize{$\pm$8.90} & 57.67 & 12.69\scriptsize{$\pm$5.21} & 67.57\scriptsize{$\pm$14.63} & 85.10 \\
VitaminC & 19.64\scriptsize{$\pm$3.80} & 83.33\scriptsize{$\pm$9.59} & 87.49 & 98.23\scriptsize{$\pm$1.05} & 67.40\scriptsize{$\pm$3.06} & 85.76& 28.97\scriptsize{$\pm$8.21} & 83.78\scriptsize{$\pm$12.12} & 93.19 \\
\midrule
Gemini$_{\texttt{1.0 - Ultra}}$ & 36.79\scriptsize{$\pm$4.23} & 59.09\scriptsize{$\pm$6.35} & - & 96.46\scriptsize{$\pm$1.13} & 90.49\scriptsize{$\pm$2.74} & - & 56.60\scriptsize{$\pm$10.15} & 81.08\scriptsize{$\pm$11.42} & - \\
\bottomrule

\end{tabular}
}
\vspace{-0.07in}
\caption{
     Per-label classification  precision and recall on \datasetname from T5-Large finetuned on different combinations of five NLI datasets, and Gemini$_{\texttt{ultra}}$ with 8-shot prompting for comparison. \texttt{ALL} denotes using the mixture of all five datasets for finetuning, and \texttt{ALL - X} denotes the leave-X-out mixture. We generally observe that all combinations of training data leads many false contradiction and false entailment in predictions. Number in parathenses shows label count in the benchmark. $\pm$ shows $95\%$ confidence interval of precision and recall, estimated via bootstrap resampling with 500 iterations.  All metrics shown are scaled by $100\times$ for visualization purposes. }
\label{tab:refnli}
\end{table*}

\paragraph{Annotation process.}
The authors went through a total of 1,143 example pairs, where one author produced the initial label and another author verified and adjudicated the label.  
On a sub-sample of 102 claims, we ask three authors to produce the label individually and we observe 0.83 Fleiss' $\kappa$ under 4-way classification, suggesting a good inter-rater agreement under the setting. In the rest of the paper, we denote the annotated set of examples as the \datasetname benchmark. 

\paragraph{Statistics.}
In \datasetname, the authors went through a total of 1,143 pairs of claim and evidence sentences, with 905 \emph{neutrals}, 66 \emph{contradictions}, 37 \emph{entailments}, and 135 \emph{ambiguous} cases.

\section{Evaluating Model's Reference Determinacy Biases}
\label{sec:results}
With \datasetname, we try to understand the effect of training datasets on the resulting NLI models' capabilities of recognizing reference determinacy. For this, we finetune a T5-large \cite{raffel2020exploring} model on different combinations of NLI datasets, and study their behaviour on \datasetname. 

\subsection{Experimental Settings}
\paragraph{Datasets.} We study a mixture of five large-scale NLI datasets: SNLI \cite{bowman-etal-2015-large} MNLI, \cite{williams-etal-2018-broad},
ANLI, \cite{nie-etal-2020-adversarial} and VitaminC \cite{schuster-etal-2021-get} and the processed NLI sentence-pair style of FEVER used in VitaminC. 

\paragraph{Training.} We initialize the model with pretrained T5-large 1.1 checkpoint using the T5x library \cite{roberts2022t5x}. We finetune the model with different combinations of the datasets, as shown in \autoref{tab:refnli}. The label set across dataset is unified to match the three-way classification on MNLI and SNLI, where each label is represented as a single token in the T5 output vocabulary space. For variations of training dataset (mixtures), we use a learning rate of $1e-4$ with the Adam optimizer \cite{Kingma2014AdamAM} and batch size of 128 during finetuning. 

\paragraph{Evaluation.} We evaluate each finetuned model on all examples in \datasetname. We report the per-label precision and recall of predicted label, which is computed by the output label token with the highest softmax probability. To account for the effect of using different classification thresholds for each label in label imbalanced setting, we additionally report the per-label area under ROC (AUROC) score over the output label probability distribution under one-label-vs-rest setting. 

We additionally evaluate Gemini$_{\texttt{ultra}}$ with 8-shot in context learning \cite{team2023gemini} as a point of comparison to contrast the behavior of finetuned NLI models with an instruction tuned large langauge model. 
\subsection{Results}
\autoref{tab:refnli} shows the classification results. We generally observe that models exhibit low precision and high recall on both contradiction and entailment predictions, suggesting the presence of many false positive predictions made on the two labels. In terms of AUROC, it's more visibly clear that models perform generally worse on recognizing contradictions compared to recognizing entailments, which echoes our observations in \cref{sec:benchmark}.

\paragraph{All training datasets show similar patterns of false contradictions and entailments.} Across all combinations of training datasets, we observe similar patterns of many false contradiction and entailment predictions, with slight variations across datasets. With respect to entailment predictions, we see almost all training configurations lead to high AUROC score (i.e. $>0.85$). However, with respect to contradictions, we observe a larger discrepency across different datasets. We observe that including SNLI and Fever(NLI) in the training mix would lead to worst performance in terms of contradiction detection.  In both leave-one-out and single dataset training settings, we observe ANLI to be the most useful dataset to include during training, especially for contradiction detection. Interestingly, ANLI (arguably) happens to be the one dataset where the reference determinacy assumption is least enforced during the annotation process, yet no definitive conclusion can ever be drawn here due to the existence of many other confounders.

On Gemini$_{\texttt{ultra}}$, we observe a much lower rate of false contradiction and entailment compared to all of the finetuned NLI models. That said, there still exists a gap between the performance on contradictions vs. entailments. For Gemini, we do not report the AUROC score as we do not have access to the output token probabilities during inference. 


\begin{table}[t]
\small
    \centering
    \begin{tabular}{c|ccc}
    \toprule
    \multirow{ 2}{*}{Model} & \multicolumn{3}{c}{$F_1$ score w.r.t each label}  \\
    & \emph{Entails} & \emph{Neutral} & \emph{Contradicts} \\
    
    \midrule
    T5-Small & 84.14 & 84.64 & 78.02 \\
    T5-Base & 88.91 & 88.42 & 82.36 \\
    T5-3B & 93.79 & 92.19 & 87.95 \\
    BERT-Tiny & 71.78 & 75.65 & 68.09 \\ 
    BERT-Base & 85.85 & 85.88 & 80.10 \\
    BERT-Large & 89.13 & 88.11 & 82.63 \\
     \bottomrule
    \end{tabular}
    \caption{Per-Label F$_1$ score of different models finetuned on MNLI and tested on MNLI validation set. We observe that model generally perform worse on contradictions compared to the other two labels.}
 \label{tab:model-perform-worse}
\end{table}

\begin{table}[t]
    \centering
\resizebox{\linewidth}{!}{
    \begin{tabular}{c|ccc}
    \toprule
    \multirow{ 2}{*}{Label} & \multicolumn{3}{c}{Metric}  \\
    & \emph{Precision}~$\uparrow$ & \emph{Recall}~$\uparrow$ & \emph{AUROC}~$\uparrow$ \\
    
    \midrule
    Entail. & 15.76 $\rightarrow$ \textbf{32.26} & \textbf{89.18} $\rightarrow$ 84.85 & 90.91 $\rightarrow$ \textbf{94.57} \\
    Neutral & \textbf{98.99} $\rightarrow$ 97.81 & 53.92 $\rightarrow$ \textbf{69.09} & 87.49 $\rightarrow$ \textbf{88.49} \\
    Contra. & 15.76 $\rightarrow$ \textbf{20.29} & \textbf{92.42} $\rightarrow$ 84.85 & 90.91 $\rightarrow$ \textbf{91.18} \\
     \bottomrule
    \end{tabular}
}
    \caption{Per-Label precision recall and AUROC of T5-large trained on the mixture of five datasets before $\rightarrow$ after training set filtering described in \cref{ssec:mitigating}}
 \label{tab:mitigating}
\end{table}

\subsection{Are Contradictions More Difficult to Learn?}
In the previous section, we observe a wide performance gap when finetuned NLI models are applied to recognize contradictions in settings where reference determinacy cannot be assumed. An additional factor here is that contradiction might be inherently a more difficult problem to learn from the training data distribution. \autoref{tab:model-perform-worse} shows an experiment where we finetune different variants of BERT \cite{devlin-etal-2019-bert} and T5 on the MNLI training set. When we evaluate the models on the MNLI dev set, we observe that the model consistently perform worse on contradiction examples. 
Here we hypothesize that the low validation performance of contradictions might be attributed to the inherent human disagreement \cite{pavlick-kwiatkowski-2019-inherent}, where the human raters tend to have more disagreements on contradictions compared to the other labels. We show and discuss evidence of this, as well as how this can be connected to the reference determinacy assumption later in \cref{ssec:human-disagree}.

\begin{table*}[t]
    \centering
\resizebox{\linewidth}{!}{
    \begin{tabular}{ccccc}
    \toprule
    \multirow{ 2}{*}{Dataset} & \multirow{ 2}{*}{Ambiguous Reference?} & \multicolumn{3}{c}{Correlation Between Human Votes ($\downarrow$)}  \\
    & & \emph{Ent.} $\leftrightarrow$ \emph{Neu.} & \emph{Ent.} $\leftrightarrow$ \emph{Con.} & \emph{Con.} $\leftrightarrow$ \emph{Neu.} \\
    
    \midrule
    \multirow{ 3}{*}{SNLI \cite{bowman-etal-2015-large}} & \textit{All} & -0.63$^{**}$ & -0.73$^{**}$ & \textbf{-0.08}$^{*}\ \ $ \\ \cmidrule{2-5}
    & No ($\tilde~53\%$) & -0.74$^{**}$ & -0.48$^{**}$ & \textbf{-0.23}$^{**}$ \\ 
    & Yes ($\tilde~47\%$) & \textbf{-0.36}$^{**}$ & -0.51$^{**}$ & -0.61$^{**}$
    \\
    \midrule
    \multirow{ 3}{*}{MNLI \cite{williams-etal-2018-broad}} & \textit{All} & -0.62$^{**}$ & -0.50$^{**}$ & \textbf{-0.37}$^{**}$ \\ \cmidrule{2-5}
    & No ($\tilde~54\%$) & -0.64$^{**}$ & -0.74$^{**}$ & \textbf{-0.03}$\ \ \ \ $
 \\
    & Yes ($\tilde~46\%$) & -0.52$^{**}$ & -0.70$^{**}$ & \textbf{-0.25}$^{**}$ \\
     \bottomrule
    \end{tabular}
    }
    \caption{To understand how reference ambiguity affects human agreement in NLI, we compute the Pearson correlation among 100 human votes per example provided in ChaosNLI \cite{nie-etal-2020-learn}. Correlation of $-1$ indicates perfect agreement among raters on the distinction between two labels, and vice versa. We randomly sample 500 examples respectively from SNLI and MNLI split of ChaosNLI and annotated whether each example contains ambiguous reference or not. (* denotes $p < 0.05$, ** denotes $p < 0.01$ for the correlation coeffecient.) }
    \label{tab:human-corr}

\end{table*}

\subsection{Mitigating the Effect of Reference Determinacy}
\label{ssec:mitigating}
To further validate that the reference determinacy assumption in the training data has an impact on downstream performance, we demonstrate that filtering out examples where reference determinacy cannot be easily determined improves the resulting model's performance on \datasetname. 

With the mixture of five training datasets, we check whether a contradiction or entailment example is likely to be affected by the reference determinacy assumption, by the simple heuristics of lexical overlap. If a hypothesis and the premise share a token-level Jaccard similarity less than or equal to $0.15$, we would discard this example from the training set, as we conjecture that it is more likely
that the example is only labeled as contradiction or entailment due to the RD assumption. We filter out such examples from the training mix, and perform a rebalance of the label distribution by random re-sampling neutral examples to match the number of contradiction or entailment examples left in the dataset.  

The evaluation results are shown in \autoref{tab:mitigating}. We see that the method generally improves the precision of entailment and contradiction predictions. We also see minor improvements across all labels in terms of AUROC. The findings here further validate our hypothesis that training with examples created with the RD assumption has a trickle-down effect on the performance of NLI models in real-world settings.

\section{Can Reference (In-)determinacy Explain Human Disagreements?}
\label{ssec:human-disagree}

Next, we study whether inherent human disagreements \cite{pavlick-kwiatkowski-2019-inherent} on NLI labels can potentially be attributed, at least in part, by the reference ambiguity between the hypothesis and premise. We conduct an experiment with the ChaosNLI dataset \cite{nie-etal-2020-learn}. ChaosNLI contains samples of the original SNLI and MNLI datasets, where each example is re-labeled by 100 different crowdsource workers. 
ChaosNLI presents an interesting case for our purpose, as the human raters were not given explicit instructions to assume reference determinacy, which was instead deferred to their own judgement. 
To understand whether and how reference ambiguity might lead to human disagreements, the authors went through 500 random samples respectively from SNLI and MNLI split of ChaosNLI, and labeled whether ambiguity exists between the hypothesis and premise, following the same annotation protocol as in \cref{ssec:benchmark}.

We compute the Pearson correlation between the number of votes each label received for each NLI example. Here, a higher correlation value between two labels (e.g., $\rightarrow 1$) indicates that humans disagree and confound the two labels more often, and vice versa. \autoref{tab:human-corr} shows the our results.

\paragraph{Humans disagree more between contradiction and neutral labels.}
Overall, we observe that human raters tend to split votes between the neutral and contradiction labels more frequently than other combinations.
Notably, on SNLI, we see a much weaker negative correlation ($r=-0.08$) between contradiction and neutral, compared to the relatively strong negative correlation between the other two label pairs. On MNLI, we observe a similar pattern, yet the gap is much smaller ($r=-0.37$ between contradiction and neutral). 
When we compare the ChaosNLI annotations against the original labels from MNLI and SNLI's five-way annotation, we observe that the change in majority label happens more often between entailment vs. neutral and contradiction vs. neutral, as shown in \autoref{fig:mnli-snli-cm} in Appendix~\ref{app:disagree}.


\begin{figure}[h]
    \centering
    \includegraphics[width=0.9\linewidth]{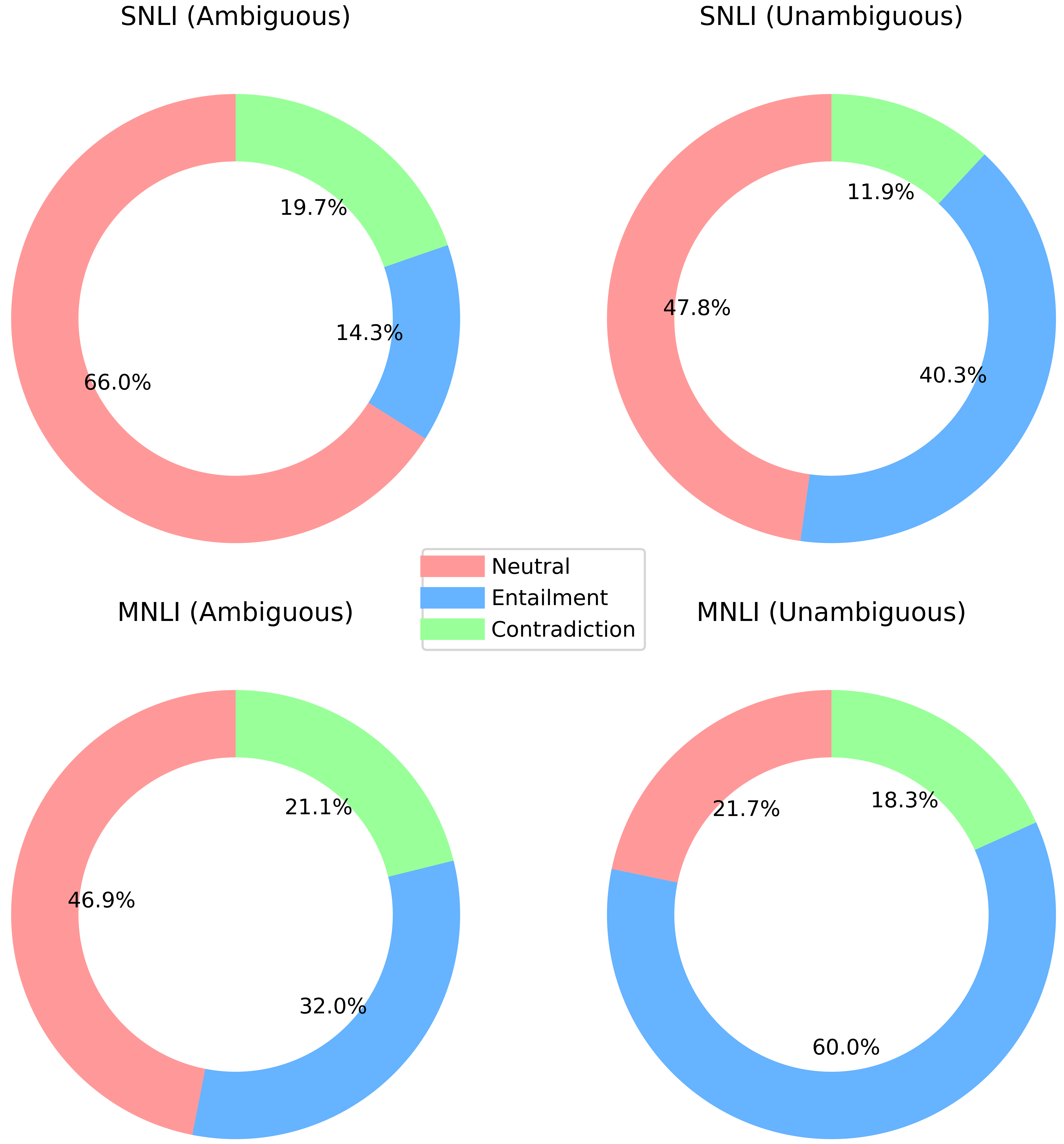}
    \caption{Distribution of the majority labels from the MNLI and SNLI split of ChaosNLI, when the reference between the hypothesis and premise is \textit{ambigious} vs. \textit{unambiguous}.}
    \label{fig:percentage}
\end{figure}

\paragraph{Human disagreements can in part be attributed to reference ambiguity.} 
To estimate the percentage of examples that exhibit disagreements due to reference determinacy, we look at how the correlation between votes on different labels changes with respect to whether reference ambiguity exists in the data. From \autoref{tab:human-corr}, we see that in both MNLI and SNLI, a large fraction of the examples exhibit the problem of reference ambiguity ($\tilde~47\%$ in SNLI, $\tilde~46\%$ in SNLI). When we compare the case between ambiguous vs. unambiguous examples, we see that on both datasets, the rater agreement between contradiction and neutral improves when we go from ambiguous to unambiguous cases, while we observe the vice versa between entailment and neutral labels. 
We observe that the change in agreement patterns are mostly due to whether the rater can safely establish reference determinacy between the hypothesis and premise. If so, then whether raters would agree on the hypothesis is contradicted by the premise is less likely to be impacted by the additional judgement of whether the two statements refer to the same context. 

In \autoref{fig:percentage}, we see how the majority label distribution shifts according to whether ambiguity exists in NLI examples. We observe that in ambiguous cases, the annotators are more likely to label an example as neutral, while in the unambiguous case, raters are more likely to judge the hypothesis as entailed or contradicted by the premise. 

The findings here echo our hypothesis that the existence of reference ambiguity in NLI examples would lead to more disagreements among annotators. This potentially suggests that human disagreement can at least in part be attributed to the reference (in-)determinacy problem, and the annotation process would have more disagreement especially when raters are not explicitly instructed to assume RD during the annotation process.



\section{Related Work}
\label{sec:related}
As ambiguity is an indispensable element in how we interpret and express language, many language understanding tasks require models to be able to recognize the resolve the ambiguity that exists in an user query \cite{xu-etal-2019-asking,zamani2020mimics,stelmakh-etal-2022-asqa,feng-etal-2023-generic,zhao2024beyond}. For instance,  \citet{min-etal-2020-ambigqa} observe that ambiguous questions might lead to different answers depending on what the user intent is, and this would lead to annotation ambiguities when raters are asked to provide a single answer for an ambiguous question. With NLI, previous studies \cite{pavlick-kwiatkowski-2019-inherent,nie-etal-2020-learn} have found that inherent human disagreements exist in NLI labels, and the disagreement usually follows instance-dependent pattern.   
This work explores the understudied problem of explaining and understanding the cause of disagreements. Being able to understand the disagreements can potentially lead to the development of better NLI systems, as 
\citet{zhou-etal-2022-distributed} and \citet{zhang-de-marneffe-2021-identifying} show the merit of modeling the uncertainty distribution of NLI labels.

Our work tries to understand the impact of annotation artifacts \cite{gururangan-etal-2018-annotation, bowman-etal-2020-new} on the downstream applicability of NLI tasks and models. In practice, researchers have found that NLI models would exploit such artifacts  \cite{poliak-etal-2018-hypothesis, mccoy-etal-2019-right}, which potentially hurts the downstream applicability. Our work is motivated by the use case of using NLI for verifying text and factual consistency \cite{schuster-etal-2021-get, schuster-etal-2022-stretching, honovich-etal-2022-true, gao-etal-2023-rarr}, and we seek to understand the limitation of NLI models in such use cases. To this end, a series of recent studies \cite{chen-etal-2023-propsegment, chen-etal-2024-sub, havaldar2025entailed} investigate alternative NLI task formulation and model architecture that incorporate additional context into NLI decisions. 

Beyond NLI and its downstream applciations, it remains to be seen whether the reference or context ambiguity problem exists in other tasks and datasets as well.  Along this line,  \citet{liu-etal-2023-afraid} designs a suite of tests that show current instruction-tuned language models often fail to respond to input ambiguity. \citet{malaviya2024contextualized} study how under-specification of context can lead to lower agreement and unreliable evaluation conclusions when doing model evaluations.  
From these findings, We conjecture that this could be due to the inherent reference ambiguity in other tasks during the instruction-tuning stage of these models. We hope to explore this thread in future work. 

\section{Conclusion}
\label{sec:conclusion}

This paper studies the impact of the reference determinacy assumption in the NLI dataset creation process. We release the \datasetname benchmark, and investigate the trickle-down effect of reference ambiguity in NLI on both the human annotators and subsequently on the NLI model training process. We hope that future NLI researchers and practitioners pay attention to this problem, especially when trying to apply NLI models in downstream use cases.

\section*{Limitations}
Our study focuses on understanding the implication of reference (in-)determinacy and its impact from a data perspective. Our modeling experiments use one fixed architecture with different mixtures of NLI datasets for training. Although it is mostly due to the fact that we want to understand the impact of using different types of NLI datasets for training, experimenting with more models could potentially eliminate model architecture as the confounder in our results. 
Although not the focus of our study, but the study could be extended and strengthened with experiments with large language models to understand the models react and respond to ambiguities in the input with the NLI task format. As we discussed at the end of \cref{sec:related}, we leave the two parts for future exploration.

\section*{Ethical Considerations}
To the best of our knowledge, our study does not introduce ethical concerns.

\bibliography{anthology,custom}

\newpage
\appendix

\section{RefNLI Annotation Guidelines}
\label{appendix:rater-guidelines}

The expert raters for RefNLI were presented with examples consisting of a premise and a hypothesis. For each example, they were given  instructions as follows.

You are to assign one of 4 labels to the example:

\begin{enumerate}
    \item[(a)] \textbf{ambiguous reference}: If the premise contains ambiguous reference, and it’s possible that with resolved reference, premise would actually support/contradict the claim. 
    \item[(n)] \textbf{neutral}: If the premise can’t support or contradict the claim in any possible way. e.g. No matter how you resolve the reference, the premise would still be irrelevant to the claim.
    \item[(c)] \textbf{contradiction}: If the claim is most likely false given the premise.
    \item[(e)] \textbf{entailment}: If premise fully supports the claim.
\end{enumerate}

If you find tricky cases, put yourself in the following scenario: Suppose an LLM generates the claim, you want to decide if we should, given the evidence, tell the user that that this claim is true, tell the user that it's false, or neither. 

The distinction between neutral and ambiguous is going to be difficult sometimes. See examples below for what we are after. If it’s truly unclear – feel free to skip the example.

\subsection*{Specific Guidelines}

\begin{enumerate}
    \item \textbf{Skip unclear claims or premises}: If you think the claim is difficult to understand, or there is too much ambiguity, skip the claim entirely.

    \item \textbf{Don’t label the claim by its truth value in the world}: If a claim says “The sky is blue”, and the premise says something completely different, label it as neutral. Don’t label such cases as entailment based on \textbf{just} your world knowledge.

    \item \textbf{World Knowledge is permitted}: You can assume commonly accepted world knowledge when interpreting the premise, e.g., basic geography and other commonsense knowledge are allowed. If needed, a web search is allowed when making the judgements. However, don’t make too many inferences.

    \item \textbf{Temporal considerations}: Ignore tense (e.g., past or present) in both the premise and claims. If the premise clearly indicates a time of an event, but the claim doesn’t, assume that the claim is uttered right after the event.

    \item \textbf{Personal surnames}: If only the surname of a person is mentioned in the premise, and there’s not enough evidence for in the premise for you to determine the last name is referring to the same entity as in the hypothesis, mark the example as ``ambiguous''

    \item \textbf{Neutral vs. Ambiguous Reference}: The distinction between the two can be difficult sometimes. The general rule is: if the premise can’t seem to support the claim no matter how you interpret the premise, then it's neutral.
\end{enumerate}

\subsection*{Some examples given in the instructions}

\newcommand{\instex}[3]{\noindent\textbf{Premise}: #1

\noindent\textbf{Hypothesis}: #2

\noindent\textbf{Label}: #3\vspace{1ex}}

\instex{Wales has a large region rich in coal deposits.}{The Ural Mountains contain about 48 species of economically valuable ores and economically valuable minerals.}{N; even if we didn't know whether the Ural Mountains are in Wales, the premise doesn’t mention anything about coal deposits, so there’s no way that the premise can support/contradict the claim.}

\instex{Wales has a large region rich in coal deposits.}{Famous for its coal, Newcastle is the largest coal exporting harbour in the world, exporting 159.9 million tonnes of coal in 2017.}{A: The prominent Newcastle is in New South Wales, Australia, but there happens to also be a small town named Newcastle in Wales}

\instex{The Predator made more than \$97 million worldwide.}{Up to March 2011, The Predator’s worldwide gross has reached \$172,543,519, making it the highest-grossing film in the franchise.}{E; if the premise mentions a time, and there’s no clear temporal marker in the claim – assume that the claim is made in the similar time frame as the premise.}

\instex{The Hunchback of Notre Dame is a Disney media franchise, commencing in 1996 with the release of "The Hunchback of Notre Dame".}{The Hunchback of Notre Dame has only ever been based off of a poem.}{Skip, since it's unclear in the hypothesis what ``based off of a poem'' means}

\section{Human Disagreements and Reference Ambiguity}
\label{app:disagree}

\begin{figure}[t]
    \centering
    \includegraphics[width=0.7\linewidth]{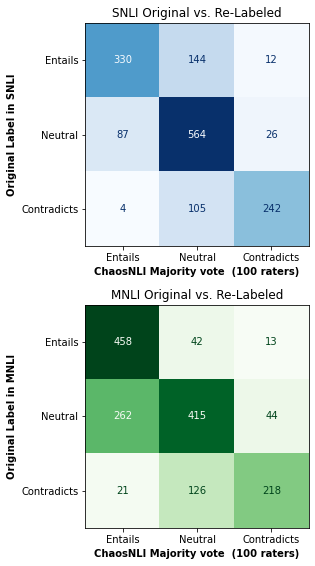}
    \caption{Confusion Matrices between majority label from the original annotation vs. ChaosNLI's re-annotation label for SNLI and MNLI examples from \citet{nie-etal-2020-learn}.}
    \label{fig:mnli-snli-cm}
    \vspace{-5pt}
\end{figure}

\autoref{fig:mnli-snli-cm} shows the confusion matrix between the majority NLI label from the ChaosNLI re-annotation vs. the original majority label from the five SNLI/MNLI annotators originally. 

\end{document}